\documentclass[conference]{IEEEtran}
\IEEEoverridecommandlockouts
\usepackage{cite}
\usepackage{amsmath,amssymb,amsfonts}
\usepackage{algorithmic}
\usepackage{graphicx}
\usepackage{textcomp}
\usepackage{tabularray}
\usepackage{xcolor}
\usepackage{hyperref}

\def\BibTeX{{\rm B\kern-.05em{\sc i\kern-.025em b}\kern-.08em
    T\kern-.1667em\lower.7ex\hbox{E}\kern-.125emX}}
\begin{document}

\title{AssemAI: Interpretable Image-Based Anomaly Detection for Manufacturing Pipelines\\
\thanks{*These authors contributed equally to this work.}}
\author{\IEEEauthorblockN{Renjith Prasad*}
\IEEEauthorblockA{\textit{Artificial Intelligence Institute} \\
\textit{College of Engineering and Computing}\\
\textit{University of South Carolina}\\
Columbia, SC 29208, USA \\
kaipplir@mailbox.sc.edu}
\and
\IEEEauthorblockN{Chathurangi Shyalika*}
\IEEEauthorblockA{\textit{Artificial Intelligence Institute} \\
\textit{College of Engineering and Computing}\\
\textit{University of South Carolina}\\
Columbia, SC 29208, USA \\
jayakodc@email.sc.edu}
\and
\IEEEauthorblockN{Fadi El Kalach}
\IEEEauthorblockA{\textit{McNair Center} \\
\textit{for Aerospace Innovation and Research}\\
\textit{Department of Mechanical Engineering}\\
\textit{College of Engineering and Computing}\\
\textit{University of South Carolina}\\
Columbia, SC 29208, USA \\
elkalach@email.sc.edu}
\and
\IEEEauthorblockN{Revathy Venkataramanan}
\IEEEauthorblockA{\textit{Artificial Intelligence Institute} \\
\textit{College of Engineering and Computing}\\
\textit{University of South Carolina}\\
Columbia, SC 29208, USA \\
revathy@email.sc.edu}
\and
\IEEEauthorblockN{Ramtin Zand}
\IEEEauthorblockA{\textit{Intelligent Circuits, Architectures} \\
\textit{and Systems Lab} \\
\textit{College of Engineering and Computing}\\
\textit{University of South Carolina}\\
Columbia, SC 29208, USA \\
ramtin@cse.sc.edu}
\and
\IEEEauthorblockN{Ramy Harik}
\IEEEauthorblockA{\textit{CU-ICAR} \\
\textit{Clemson University}\\
\textit{College of Engineering and Computing}\\
Greenville, SC 29607, USA \\
harik@clemson.edu}
\and
\IEEEauthorblockN{Amit Sheth}
\IEEEauthorblockA{\textit{Artificial Intelligence Institute} \\
\textit{College of Engineering and Computing}\\
\textit{University of South Carolina}\\
Columbia, SC 29208, USA \\
amit@sc.edu}
}
\maketitle

\vspace{-4 mm}
\begin{abstract}
Anomaly detection in manufacturing pipelines remains a critical challenge,  intensified by the complexity and variability of industrial environments.  This paper introduces AssemAI, an interpretable image-based anomaly detection system tailored for smart manufacturing pipelines. Utilizing a curated image dataset from an industry-focused rocket assembly pipeline, we address the challenge of imbalanced image data and demonstrate the importance of image-based methods in anomaly detection. Our primary contributions include deriving an image dataset, fine-tuning an object detection model YOLO-FF, and implementing a custom anomaly detection model for assembly pipelines. The proposed approach leverages domain knowledge in data preparation, model development and reasoning. We implement several anomaly detection models on the derived image dataset, including a Convolutional Neural Network, Vision Transformer (ViT), and pre-trained versions of these models. Additionally, we incorporate explainability techniques at both user and model levels, utilizing ontology for user-level explanations and SCORE-CAM for in-depth feature and model analysis. Finally, the best-performing anomaly detection model and YOLO-FF are deployed in a real-time setting. Our results include ablation studies on the baselines and a comprehensive evaluation of the proposed system. This work highlights the broader impact of advanced image-based anomaly detection in enhancing the reliability and efficiency of smart manufacturing processes. \noindent The image dataset, codes to reproduce the results and additional experiments are available at   \url{https://github.com/renjithk4/AssemAI}.

\end{abstract}


\section{Introduction}
The evolution of manufacturing has been driven by distinct technological milestones. Initially, mechanization marked the first industrial revolution, followed by the mass production techniques of the second revolution. The third revolution introduced automation and computerization, significantly enhancing operational efficiency. 
The modern smart manufacturing paradigm emphasizes the utilization of data and advanced analytics to inform decision-making, thereby optimizing productivity and efficiency \cite{anumbe2022primer}. The incorporation of the Internet of Things (IoT), Artificial Intelligence (AI) and other advanced technologies plays a pivotal role in this transformation, revolutionizing manufacturing processes and systems \cite{tao2018data}.

Anomaly detection is an essential methodology in manufacturing to identify deviations from normal production processes, which can indicate potential issues such as equipment failures, defects, or inefficiencies. Anomaly detection helps ensure product quality, reduce downtime, and optimize production processes \cite{wu2018machine} \cite{Pang_2021}. By monitoring various process parameters and product characteristics, manufacturers can detect anomalies early and prevent costly production disruptions \cite{chandola2009anomaly}.

Image data are increasingly being leveraged in manufacturing systems due to advancements in computer vision and the availability of high-resolution cameras. In modern manufacturing facilities, image data are utilized for a wide range of applications, including quality inspection \cite{babic2021image, sundaram2023artificial}, predictive maintenance \cite{drakaki2022machine, kiangala2020effective}, process monitoring \cite{du2018fault} and process optimization \cite{yang2015online}. The integration of image data allows for real-time monitoring and anomaly detection, significantly improving the ability to identify defects and streamline production processes. Specifically, image-based anomaly detection has gained prominence, leveraging visual data to identify defects or irregularities in products. Automated visual inspection systems use image data to detect surface defects, assembly errors and deviations from design specifications\cite{article}, providing real-time feedback essential for maintaining high production standards and minimizing defects\cite{Pang_2021}.

Despite the advancements in image-based anomaly detection, several challenges persist as follows: 
(i) In visual data, the distinctions between normal and anomalous conditions are often subtle, making it difficult for the model to differentiate these anomalies from typical image variations
(ii) The need for large and diverse datasets for training can be resource-intensive, both in terms of data collection and annotation \cite{chandola2009anomaly,googleCVPR2023Tutorial}.
(iii) Generalization issues that arise when a model trained in one manufacturing environment fails to adapt to others due to differences in lighting, camera angles, or product types \cite{yao2022generalizableindustrialvisualanomaly}.
(iv) Difficulty in interpreting detection results, hindering the ability to provide actionable insights for process improvement \cite{tian2021review}. To address these challenges, this work presents the following contributions:

\begin{enumerate}
      \item A new image dataset for anomaly detection in manufacturing, derived using a zero-shot object detection model OWL-ViT.
    \item A novel anomaly detection model that leverages the pre-trained EfficientNet architecture \cite{tan2019efficientnet}, fine-tuned on the derived dataset for enhanced anomaly detection in assembly pipelines.
\item Integration of user-level explainability using process ontology and model-level explainability using SCORE-CAM for salient feature and model analysis, 

\end{enumerate}

\begin{figure*}[!t]
  \centering
   \includegraphics[width=0.9\linewidth]{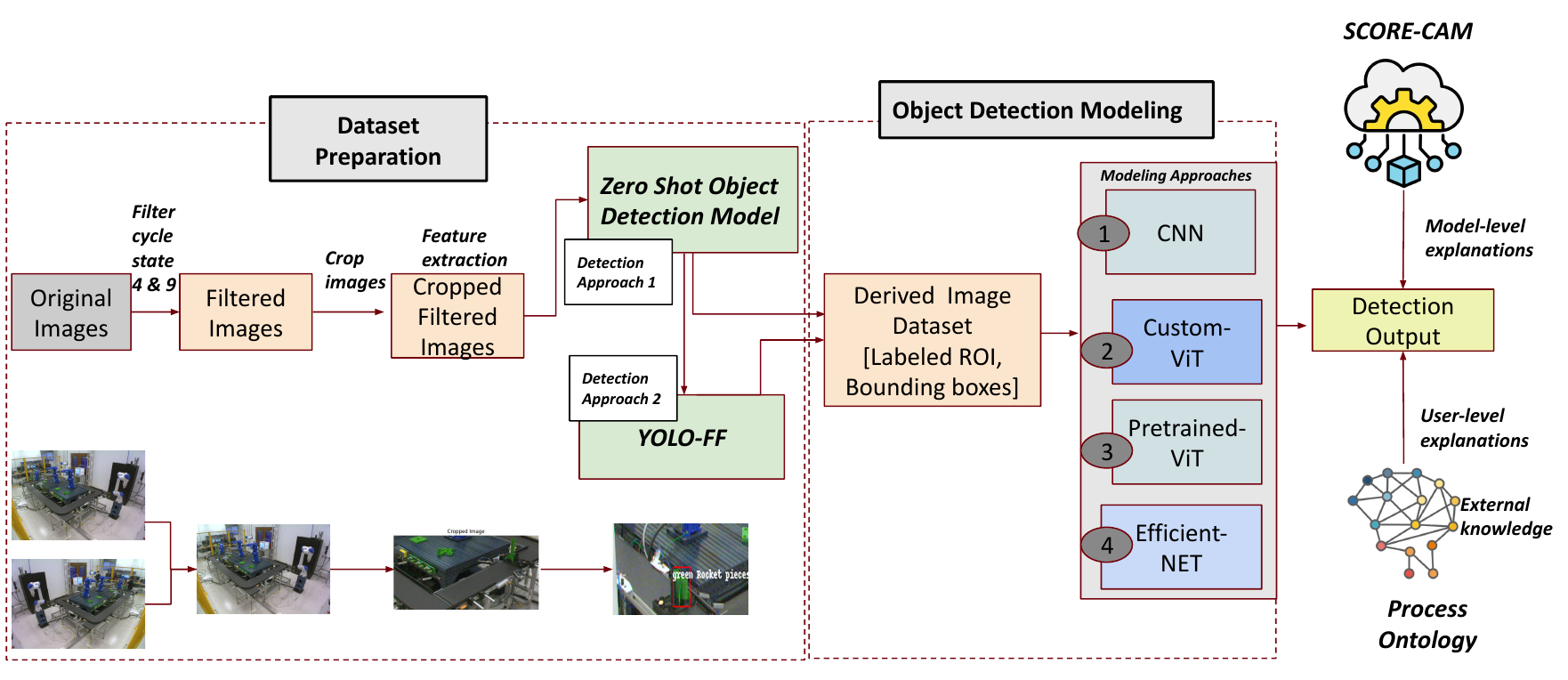}
    \vspace{-2mm}
  \caption{Overall Architecture of AssemAI. 
The figure illustrates the AssemAI pipeline, beginning with dataset preparation, which includes filtering and cropping images. Next, object detection is performed using zero-shot detection and a fine-tuned model (YOLO-FF). This is followed by anomaly detection using CNN, Custom-ViT, Pre-trained-ViT, and EfficientNet. The detection output is explained using SCORE-CAM for model-level explanations and process ontology for user-level explanations.}

  \label{fig:overall_architecture.pdf}
  \vspace{-6mm}
\end{figure*}

\vspace{-2mm}


\section{Literature Review}


\subsection{Object Detection Models}

Object detection has advanced significantly, with various models enhancing both accuracy and efficiency.
Girshick et al. \cite{girshick2014rich} introduced R-CNN, which applies CNNs to region proposals for object classification and bounding box refinement. Faster R-CNN \cite{ren2015faster} built on R-CNN by integrating a Region Proposal Network (RPN) for faster detection. The YOLO models, from YOLOv1 to YOLOv9, transformed object detection by formulating it as a single regression problem to predict bounding boxes and class probabilities directly \cite{wang2024yolov9learningwantlearn}. ShuffleNet \cite{zhang2018shufflenet} and MobileNet \cite{howard2017mobilenets} are lightweight models designed for efficiency on mobile devices. SqueezeNet \cite{iandola2016squeezenet} aimed to reduce model size while maintaining accuracy with its Fire module. Lastly, the Swin Transformer \cite{liu2021swin} introduced hierarchical transformers with shifted windows, setting new benchmarks for object detection and vision tasks. Despite these advancements, limitations such as the need for extensive labeled data and computational resources persist. Our work addresses these limitations by focusing on lightweight and efficient models tailored for anomaly detection in manufacturing settings, improving both detection speed and accuracy while maintaining interpretability.

\subsection{Zero-Shot Object Detection}

Zero-shot object detection (ZSD) recognizes objects without labeled training data but faces challenges like semantic noise and class imbalance. Foundational methods by Bansal et al. \cite{10.1007/978-3-030-01246-5_24} and Rahman et al. \cite{rahman2018zero} struggled with noise. Gupta et al. \cite{Gupta_2020_WACV} and Zheng et al. \cite{Zheng_2021} improved ZSD with symmetric mapping and cascade stages, but issues persisted. Hayat et al. \cite{hayat2020synthesizing} and Li et al. \cite{li2020context} used ResNet, KNN, and contextual information but faced noise and ambiguity. Zhu et al. \cite{zhu2020dont} and Hayat et al. \cite{hayat2020generative} advanced ZSD with Don't Even Look Once (DELO) model and Generative-ZSD. Liu et al. \cite{liu2021contrastzsd} and Li et al. \cite{li2023semantics} proposed contrastive learning and a semantics-aware framework. Our work focuses on specialized domains for precise anomaly detection.
\subsection{Anomaly detection}
Anomaly detection has seen significant advancements across diverse domains like networking \cite{kim2006image}, smart agriculture \cite{mendoza2024convolutional}, healthcare \cite{alloqmani2021deep}, manufacturing \cite{xie2024iad,maggipinto2019deep,jiang2019machine,tan2019encoder,kim2023self,bougaham2024composite,10129339}. Haselmann et al. \cite{haselmann2018anomaly} present an unsupervised one-class learning method using a deep CNN for surface inspection, outperforming other methods on decorated plastic parts. Xie et al. \cite{xie2024iad} propose a uniform benchmark for assessing image anomaly detection (IAD) algorithms in industrial settings. Maggipinto et al. \cite{maggipinto2019deep} use convolutional autoencoders for monitoring semiconductor manufacturing, enhancing effectiveness and scalability. Jiang et al. \cite{jiang2019machine} introduce YOLOv3 for balanced datasets and Fast-AnoGAN for unbalanced datasets in industrial production. Tan et al. \cite{tan2019encoder} use an encoder-decoder for anomaly detection in sequential sensor data. Kim et al. \cite{kim2023self} propose a self-supervised method using Gramian angular field and StyleGAN for time-series data. Bougaham et al. \cite{bougaham2024composite} demonstrate a three-step deep learning approach for Printed Circuit Board Assembly (PCBA) images, achieving high accuracy. Despite advancements, publicly available datasets for assembly processes are scarce, and existing methods lack interpretability and explainability for domain-specific insights.

\section{Proposed Methodology}
Figure~\ref{fig:overall_architecture.pdf} shows the overall architecture of the proposed method. 
The AssemAI pipeline begins with dataset preparation, where images are filtered and cropped to focus on relevant features. We then use a zero-shot object detection model to get the region of interest for the rocket in the image, deriving a dataset with masked regions of interest and bounding box details in a CSV file. A YOLO model is then fine-tuned on these images and bounding boxes. Subsequently, this derived dataset is used for anomaly detection experiments with several architectures, including CNN, Custom ViT, Pre-trained ViT, and EfficientNet.

\begin{table}[!htb]
\centering
\caption{Artificats of the original images in FF Multimodal dataset}
\label{tab:artifacts}
\begin{tabular}{|l|l|} 
\hline
\textbf{Dataset Artifact} & \textbf{Statistic}  \\ 
\hline
Rarity Percentage         & 35.73\%             \\ 
\hline
Frequency                 & 0.367 Hz            \\ 
\hline
Data collection period    & 30 hours            \\ 
\hline
Total image count         & 332002              \\ 
\hline
Original image size       & 720px*1080px          \\ 
\hline
Types of anomalies        & 7                   \\ 
\hline
Types of classes          & 8                   \\
\hline
\end{tabular}
\vspace{-2mm}
\end{table}

The detection output is then explained using SCORE-CAM for model-level explanations and integrated with process ontology for user-level explanations, providing technical and domain-specific insights.

\subsection{Future Factories Multimodal Dataset}
We use the Future Factories (FF) dataset \cite{harik2024analog} curated and publicly released by the Future Factories lab at the McNair Aerospace Research Center, University of South Carolina. The dataset consists of measurements from a simulation of a rocket assembly pipeline, which adheres to industrial standards in deploying actuators, control mechanisms and transducers\footnote{Refer to the images available in the GitHub repository for further details.}.
The dataset comprises two versions; analog and multimodal dataset, for which we use the images included in the multimodal dataset in our study. Table \ref{tab:artifacts} shows the statistics of the original images in the FF multimodal dataset.

\subsection{Image filtering}
\begin{figure}[!htb]
  \centering
   \includegraphics[width=\linewidth]{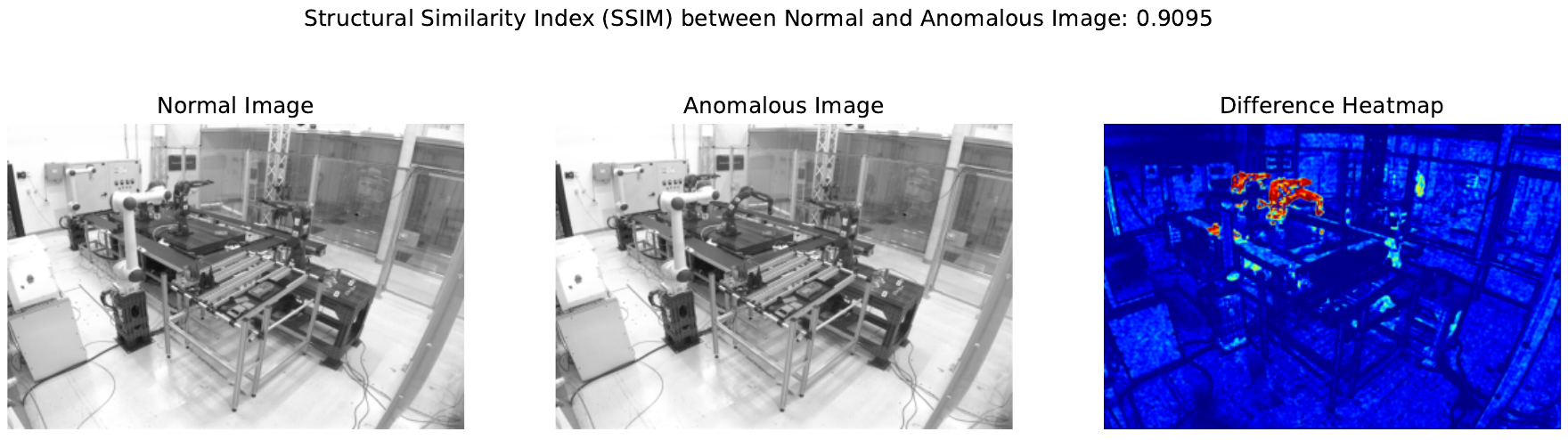}
  \caption{Structural Similarity between Normal and Anomalous Images}
  \label{fig:comparison_results_cyclestate8-cropped}
\end{figure}

The rocket assembly process at the FF lab is divided into 21 distinct cycle states. Information about these cell cycle states is not directly available in the multimodal dataset and had to be extracted from the analog dataset using a mapping function provided by domain experts. By calculating the Structural Similarity Index (SSIM) between the normal and anomalous images as shown in Figure \ref{fig:comparison_results_cyclestate8-cropped}, we observed that the images are structurally very similar across most cycle states. This indicates that the differences between the images are subtle and localized, which suggests that further analysis is needed to filter and figure out the region of interest. Among these cell cycle states, the rocket and its parts are visible only in two specific cycle states due to the spatial location of robots and other machinery and the location of the cameras and camera angles. We focus on filtering cycle four and a section of cycle nine based on domain knowledge and observational insights for enhanced image understanding. Table \ref{tab:fil_artifacts} and \ref{tab:fil_anomalies} show the filtered dataset and its anomaly statistics, respectively.

\begin{table}[!htb]
\centering
\caption{Artificats of the filtered images in FF Multimodal dataset}
\label{tab:fil_artifacts}
\begin{tabular}{|l|l|} 
\hline
\textbf{Dataset Artifact}                                       & \textbf{Statistic}                                             \\ 
\hline
Total image count                                               & 15594                                                          \\ 
\hline
Train image count                                               & 12475                                                          \\ 
\hline
Test image count                                                & 3119                                                           \\ 
\hline
\begin{tabular}[c]{@{}l@{}}Filtered image \\sizes\end{tabular} & \begin{tabular}[c]{@{}l@{}}1:200px*70px (cycle state 4) \\2:400px*205px (cycle state 9)\end{tabular}  \\
\hline
\end{tabular}
 \vspace{-2mm}
\end{table}

\begin{table}[!htb]
\centering
\caption{Anomaly types in filtered Multimodal FF dataset}
\label{tab:fil_anomalies}
\begin{tblr}{
  hlines,
  vlines,
}
\textbf{Anomaly type}        & {\textbf{Train }\\\textbf{image }\\\textbf{count}} & {\textbf{Test }\\\textbf{image }\\\textbf{count}} & {\textbf{Total count }\\\textbf{by anomaly}} & \textbf{Percentage} \\
No Anomaly                   & 8006                                               & 2016                                              & 10022                                        & 64.26\%             \\
NoNose                       & 872                                                & 238                                               & 1110                                         & 7.1\%               \\
{NoNose,\\NoBody2}           & 1222                                               & 308                                               & 1530                                         & 9.8\%               \\
{NoNose,\\NoBody2,\\NoBody1} & 1310                                               & 310                                               & 1620                                         & 10.38\%             \\
NoBody1                      & 1065                                               & 247                                               & 1312                                         & 8.4\%               \\
{Total image \\count}        & 12475                                              & 3119                                              & 15594                                        &                     
\end{tblr}
\vspace{-4mm}
\end{table}

\subsection{Image Cropping}

Our dataset comprises images captured at various stages of the manufacturing cycle. In each cycle, the rocket's position remains consistent within each respective state. Using this domain knowledge, we defined a cropping strategy by defining a bounding box for each cycle state to remove the background from all images. This consistent positioning of the rockets across cycles enables us to crop the images accurately, ensuring that only the relevant parts of the images are retained.

\begin{figure}[!htb]
  \centering
   \includegraphics[width=0.8\linewidth]{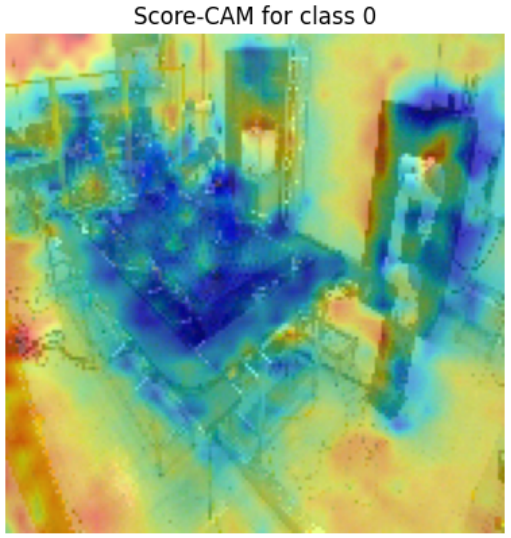}
  \caption{ Score-CAM visualization shows the model mistakenly focusing on background elements, highlighted in red as the most important regions, rather than the assembly pipeline marked in blue. This emphasizes the need for cropping and object detection to improve accuracy by isolating the relevant parts of the image.}
  \label{fig:ResultofSCORE-CAM}
  \vspace{-4mm}
\end{figure}

The suboptimal results obtained with uncropped images necessitated cropping of images. Specifically, our models performed poorly, with significant misclassification rates. We employed the explainability model SCORE-CAM \cite{Wang2019ScoreCAMSV} to understand the underlying issues. The insights provided by SCORE-CAM revealed that the models prioritized images' backgrounds rather than the rockets themselves, as shown in Figure \ref{fig:ResultofSCORE-CAM}. Therefore, images were cropped to remove the backgroun. This image transformation improved the model performance, leading to better performance. As the rocket is the most important feature, the images are cropped accordingly, which resulted in a 200x70 pixel image for cycle state four and 400x205 pixels for cycle state nine.

\subsection{Text Guided Zero-Shot Object Detection}
We then incorporate a zero-shot object detection model to enhance the accuracy and robustness of our anomaly detection pipeline, which targets the detection of rockets and their parts. Specifically, we employ OWL-ViT \cite{minderer2022simpleopenvocabularyobjectdetection}, a model that leverages multimodal representations to perform open-vocabulary detection. This approach allows for the detection of objects based on free-text prompts, facilitating a flexible and powerful detection mechanism. OWL-ViT integrates CLIP (Contrastive Language-Image Pretraining) with lightweight object classification and localization heads. This integration enables the model to handle open-vocabulary detection by embedding free-text queries through CLIP's text encoder. These are then utilized as inputs for the object classification and localization heads. The ViT processes image patches as inputs, associating them with their corresponding textual descriptions.

\subsection{
Fine-tuning Object Detection Model with YOLOv9}
To further enhance the object detection pipeline for rocket assembly, we fine-tune a YOLOv9 object detection model \cite{wang2024yolov9learningwantlearn} using the dataset derived from OWL-ViT, which we refer to as YOLO-FF. While zero-shot models like OWL-ViT offer significant flexibility and adaptability, a fine-tuned model provides specific advantages as follows:

\begin{itemize}
    \item Fine-tuning YOLOv9 using the derived dataset ensures the model parameters are optimized for the specific task, facilitating better accuracy and precision.

    \item YOLOv9 is optimized for high-speed inference, making it ideal for real-time applications, whereas models like OWL-ViT typically require more computation and have longer inference times.

\end{itemize}

\section{Problem Formulation}

Consider a dataset \( \mathcal{D} \) comprising images captured at various stages of the rocket assembly process. The manufacturing process is divided into cycles, each consisting of 21 distinct states. Let \( \mathcal{C} = \{C_1, C_2, \ldots, C_{n}\} \) denote the set of manufacturing cycles, where each cycle \( C_j \) for \( j = 1, 2, \ldots, n \) represents the complete assembly of one rocket. Each cycle \( C_j \) is divided into 21 states, represented by the set \( \mathcal{S} = \{S_1, S_2, \ldots, S_{21}\} \). For our anomaly detection task, we focus on images corresponding to cycle states \( S_4 \) and \( S_9 \), where rocket parts are most likely to be visible. Let \( \mathcal{I} = \{I_{j,s} \mid j = 1, 2, \ldots, n, \, s \in \{4, 9\}\} \) denote the set of images captured during these specific cycle states. Each image \( I_{j,s} \) is associated with a label \( y_{j,s} \in \mathcal{L} \), where \( \mathcal{L} = \{\text{normal}, \text{anomaly}_1, \text{anomaly}_2, \ldots, \text{anomaly}_k\} \). Additionally, let \( B_{j,s} = (x_{\min}, y_{\min}, x_{\max}, y_{\max}) \) denote the bounding box for image \( I_{j,s} \), obtained via object detection techniques to focus on the relevant part of the image. We define the dataset as in equation 1.
\begin{equation}
\mathcal{D} = \{(I_{j,s}, y_{j,s}, B_{j,s}) \mid j = 1, 2, \ldots, n, \, s \in \{4, 9\}\}
\end{equation}
\subsection{Task Description}
The task is to develop a model \( f: \mathcal{I} \to \mathcal{L} \) that can accurately classify each image \( I_{j,s} \) into one of the predefined categories in \( \mathcal{L} \). The classification involves the following steps:
\begin{enumerate}
    \item Filtering the dataset to include only images from cycle states \( S_4 \) and \( S_9 \).
    \item Cropping each image \( I_{j,s} \) using its bounding box \( B_{j,s} \), which is mapped to its specific state. This effectively isolates the regions of interest while removing background noise.
    \item Detecting objects in the cropped images using an object detection model. This will generate the region of interest (i.e., rocket parts) and their bounding boxes.
    \item Classifying anomalies from the detected objects using an anomaly detection model.
\end{enumerate}

The main goal is to detect anomalies in the images by using bounding boxes to help the model focus on the important parts. To address class imbalance, we use a weighted cross-entropy loss, where higher weights are assigned to classes with fewer images. We aim to minimize the weighted classification error defined in equation 2:
\vspace{-2mm}
\begin{equation}
\min_{\theta} \frac{1}{n} \sum_{j=1}^{n} \sum_{s \in \{4, 9\}} w_{y_{j,s}} \mathcal{CE}(f(I_{j,s}; \theta), y_{j,s})
\end{equation}
where \( \theta \) represents the model parameters, \( \mathcal{CE} \) is the cross-entropy loss function, and \( w_{y_{j,s}} \) is the weight assigned to the class \( y_{j,s} \).
By focusing on specific cycle states and using object detection to preprocess the images, we aim to improve the accuracy and reliability of our anomaly detection system. 

\section{Experiments}

We outline the following experimental setup to evaluate the AssemAI's overall performance and the contribution of each sub-module.

\subsection{Experiments with YOLO-FF Model}
The YOLOv9 model is fine-tuned on the derived labeled image dataset resulting from the OWL-ViT model. It is trained for 50 epochs. During training, the model achieved an accuracy of approximately 99\% on the validation dataset.
\subsection{Common Hyperparameters and Training Setup}
The derived image dataset is split into training (80\%) and testing (20\%) sets. We used Cross-Entropy Loss and the Adam optimizer \cite{kingma2014adam} for baselines and the proposed models, tuning hyperparameters like epochs, batch size and learning rate. The best model is saved based on validation accuracy. The dataset preprocessing involved resizing images to the required input dimensions (224×224 pixels), normalizing ([0.485,0.456,0.406]) and standardizing ([0.229,0.224,0.225]) according to the specific requirements of each model.

\subsection{Baselines}

\subsubsection{Custom CNN}

The Simple CNN model architecture includes two convolutional layers (32 and 64 filters respectively), each followed by a max-pooling layer. After flattening, the output is passed through a fully connected layer with 512 neurons, then to the final layer corresponding to the number of classes (5).

\subsubsection{Custom ViT}

The Vision Transformer (ViT) model \cite{dosovitskiy2020image} is implemented from scratch, starting with a patch embedding layer of size 16 that splits the input image into non-overlapping patches, each embedded into a high-dimensional space of size 768. This is followed by a series of transformer blocks for learning spatial relationships within the image patches. Each transformer block consists of a multi-head self-attention mechanism and a feed-forward neural network, with layer normalization and residual connections applied at each step. The final classification head maps the output to the desired number of classes (5).

\subsubsection{Pre-trained ViT}

We utilize the pre-trained Vision Transformer (ViT) model, specifically google/vit-base-patch16-224. Images are preprocessed by dividing each image into a sequence of fixed-size non-overlapping patches, then linearly embedded. A [CLS] token is added to represent the entire image, facilitating classification. Absolute position embeddings are incorporated and the resulting sequence of vectors is fed to the standard Transformer encoder. The ViTImageProcessor resizes and normalizes images to the required 224x224 resolution. The model is loaded with half-precision (torch.float16). The training utilized PyTorch Lightning’s Trainer, with a weighted Cross-Entropy loss function, and the AdamW optimizer with a learning rate scheduler. The best checkpoint is saved by monitoring the validation loss.

\begin{figure}[!htb]
  \centering
   \includegraphics[width=\linewidth]{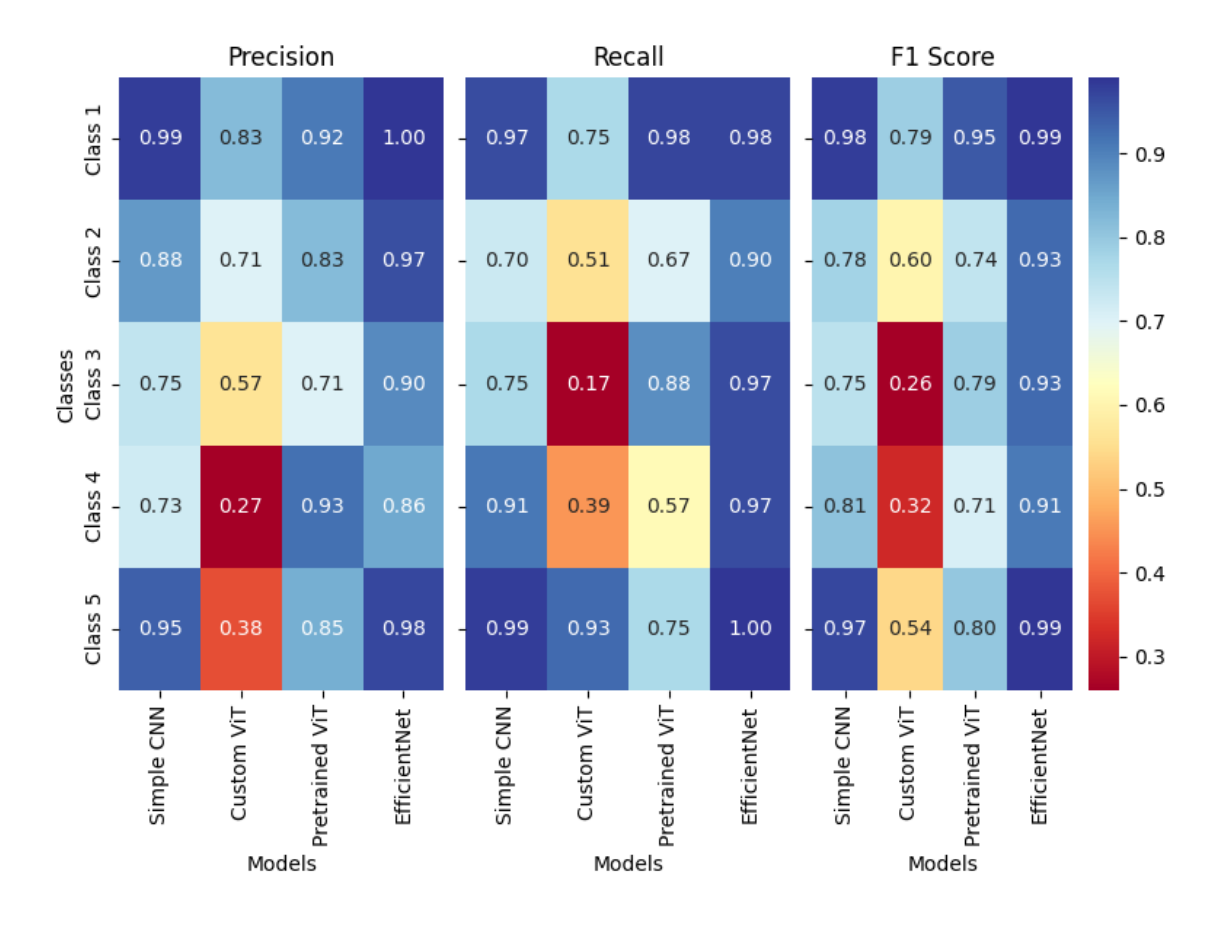}
  \caption{Experimental results with OWL-ViT. Class 1-5 denotes anomaly types: [No Anomaly], [NoNose], [NoNose,NoBody2], [NoNose,NoBody2,NoBody1] and [NoBody1] respectively}
  \label{fig:results_1}
   \vspace{-2mm}
\end{figure}

\subsection{Proposed Modeling Approach}

We implement an EfficientNet-B0 model \cite{tan2019efficientnet} for image classification tasks. The EfficientNet-B0 architecture is selected for its state-of-the-art performance and efficiency in image classification. It leverages a compound scaling method to optimize model depth, width and resolution, achieving high accuracy with fewer parameters compared to traditional networks. The model is pre-trained on ImageNet and adapted for our task by modifying the final classification layer to match the number of classes in our dataset (5 classes). 

\section{Results}
Table \ref{tab:results} summarizes the results of our experiments on the test set and the ablation studies across different baselines. We evaluate the performance using four metrics: weighted averages of precision, recall, F1-score and accuracy. The weighted averages are calculated based on the four types of anomaly classes and the normal class. The EfficientNet model achieves 96\% overall accuracy and 95\% of precision, 96\% of recall, 95\% of F1 score, respectively. Figure \ref{fig:results_1} depicts the performance of detecting various anomaly types and the normal class across various modeling approaches with the YOLO-FF object detection approach. It can be observed that among all the models, EfficientNet gives the best results in detecting all the types of classes.


\begin{table}
\centering
\scriptsize
\caption{Experimental Results of AssemAI}
\label{tab:results}
\begin{tabular}{|c|c|c|c|c|c|} 
\hline
\textbf{Model}                                                                   & \textbf{WP}                                                              & \textbf{WR}                                                              & \textbf{WF1}                                                             & \textbf{Accuracy}                                                        & \textbf{Support*}  \\ 
\hline

\begin{tabular}[c]{@{}c@{}}Simple CNN*\end{tabular}                     & \begin{tabular}[c]{@{}c@{}}82.00±\\1.00\%\end{tabular}                   & \begin{tabular}[c]{@{}c@{}}81.00±\\1.00\%\end{tabular}                   & \begin{tabular}[c]{@{}c@{}}83.00±\\1.00\%\end{tabular}                   & \begin{tabular}[c]{@{}c@{}}82.05±\\1.00\%\end{tabular}                   & 3119               \\ 
\hline
\begin{tabular}[c]{@{}c@{}}Custom ViT*\end{tabular}                     & \begin{tabular}[c]{@{}c@{}}85.00±\\1.00\%\end{tabular}                   & \begin{tabular}[c]{@{}c@{}}86.00±\\0.05\%\end{tabular}                   & \begin{tabular}[c]{@{}c@{}}83.00±\\1.00\%\end{tabular}                   & \begin{tabular}[c]{@{}c@{}}84.10±\\1.00\%\end{tabular}                   & 3119               \\ 
\hline
\begin{tabular}[c]{@{}c@{}}Pre-trained ViT* \end{tabular}                & \begin{tabular}[c]{@{}c@{}}88.50±\\0.50\%\end{tabular}                   & \begin{tabular}[c]{@{}c@{}}87.50±\\0.50\%\end{tabular}                   & \begin{tabular}[c]{@{}c@{}}87.50±\\0.50\%\end{tabular}                   & \begin{tabular}[c]{@{}c@{}}88.50±\\0.50\%\end{tabular}                   & 3119               \\ 
\hline
\begin{tabular}[c]{@{}c@{}}EfficientNet with \\original images\end{tabular}      & \begin{tabular}[c]{@{}c@{}}62.50±\\0.50\%\end{tabular}                   & \begin{tabular}[c]{@{}c@{}}60.50±\\0.50\%\end{tabular}                   & \begin{tabular}[c]{@{}c@{}}61.50±\\0.50\%\end{tabular}                   & \begin{tabular}[c]{@{}c@{}}61.50±\\0.50\%\end{tabular}                   & 332002             \\ 
\hline
\begin{tabular}[c]{@{}c@{}}EfficientNet with \\filtered images\end{tabular}      & \begin{tabular}[c]{@{}c@{}}70.50±\\0.50\%\end{tabular}                   & \begin{tabular}[c]{@{}c@{}}72.50±\\0.50\%\end{tabular}                   & \begin{tabular}[c]{@{}c@{}}73.50±\\0.50\%\end{tabular}                   & \begin{tabular}[c]{@{}c@{}}72.50±\\0.50\%\end{tabular}                   & 3119               \\ 
\hline
\begin{tabular}[c]{@{}c@{}}\textbf{EfficientNet*}\end{tabular} & \begin{tabular}[c]{@{}c@{}}\textbf{95.00±}\\\textbf{1.00\%}\end{tabular} & \begin{tabular}[c]{@{}c@{}}\textbf{96.00±}\\\textbf{1.00\%}\end{tabular} & \begin{tabular}[c]{@{}c@{}}\textbf{95.00±}\\\textbf{1.00\%}\end{tabular} & \begin{tabular}[c]{@{}c@{}}\textbf{96.00±}\\\textbf{1.00\%}\end{tabular} & \textbf{3119}      \\
\hline
\end{tabular}\\ 
\footnotesize{Models marked in * are experimented with the derived image dataset. EfficientNet model is experimented for original images (includes all cycle states) and for filtered images (includes cycle states four and nine). Bold indicates the best performance.}
\end{table}

\subsection{Additional experiments on Explainability}
\subsubsection{User level explainability}
In this work, we employ process ontology designed and developed for the Future Factories Rocket assembly line. In contrast to conventional ontologies, process ontology not only captures the definition of sensors but also captures the procedural nature of the assembly process. This aids in understanding the involvement of sensors and types of equipment at a given point in the assembly process. The Future Factories assembly is divided into 21 cycle states, which form the basis of the ontology construction process. The specific features of process ontology are as follows: (i) consists of definition and item specification of sensors and types of equipment, (ii) relationship between the sensors and types of equipment, (iii) function and involvement of each sensor and robot with respect to the cycle states (iv) expected (or anomalous) values of sensor variables in with respect to each cycle state (v) type of anomaly that could be associated with each cycle state (iv) sensor values and other knowledge can be dynamically updated as per the change in experiment set up. Capturing the procedural nature of the assembly process aids in understanding the contribution of sensors in anomalies. For example, if the initial stage of the assembly line is being analyzed for anomaly, it can be understood from the ontology that Robot-4 and its corresponding sensors do not contribute to this anomaly.

\begin{figure}[!htb]
  \centering
   \includegraphics[width=\linewidth]{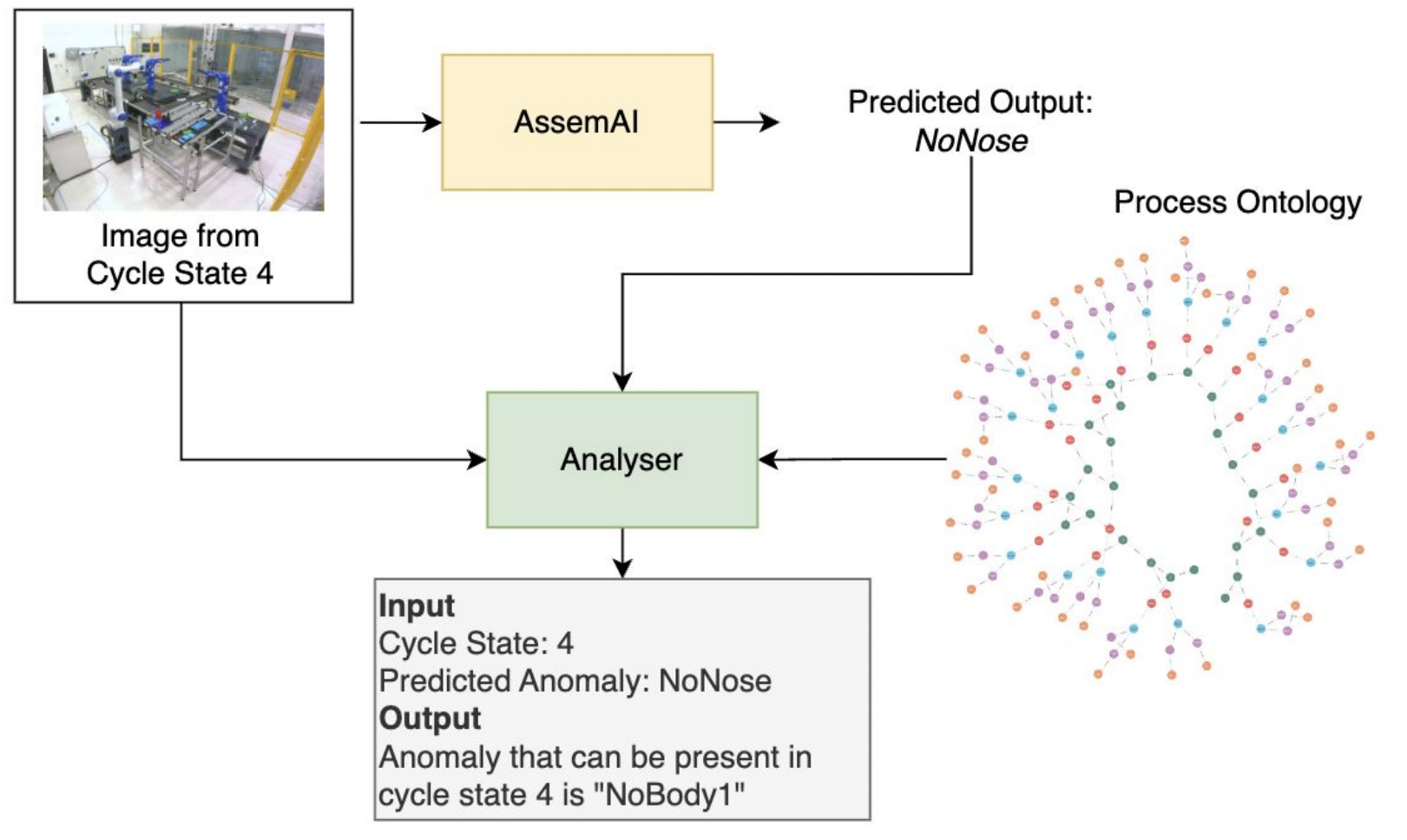}
  \caption{Verification of the explanations through process ontology}
  \label{fig:verification}
\end{figure}

The goal of process ontology is to explain and assess the output from the models. Given an image, if the model detects an anomaly, the expected values of the sensors present in that image can be obtained and provided to the user. On the other hand, process ontology can also verify if the model predictions are correct in certain cases (Figure \ref{fig:verification}). Since each image is associated with a timestamp which in turn can be mapped to the cycle state, the predicted output from the model can be verified using anomaly types defined in the process ontology. Figure \ref{fig:verification} illustrates that the input image is associated with cycle state 4. The model predicted the image to be anomalous with type \textit{NoNose}. This is an incorrect prediction as per the ontology as \textit{NoNose} anomaly can happen only from cycle stage 8 onwards. Using this knowledge and verifying the outputs of model predictions, it is found that the model incorrectly predicted \textit{NoBody2} anomaly 51 out of 801 times and \textit{NoNose} anomaly 106 out of 1145 times. The ontology also acts as an additional layer that catches the misclassification of the model, enhancing the robustness of the proposed anomaly detection pipeline.

\begin{figure}[!htb]
  \centering
   \includegraphics[width=\linewidth]{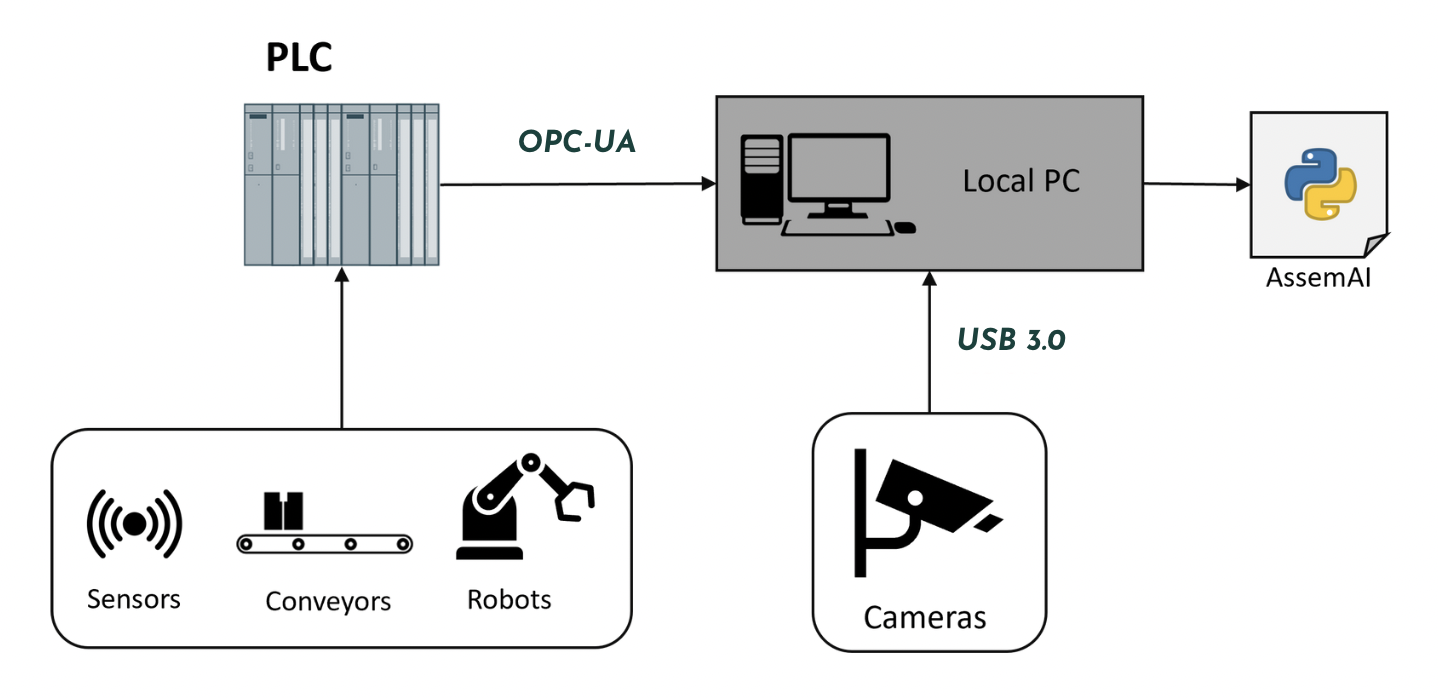}
  \caption{Deployment Setup of AssemAI}
  \label{fig:DepArch}
\end{figure}

\begin{figure}[!htb]
  \centering
   \includegraphics[width=\linewidth]{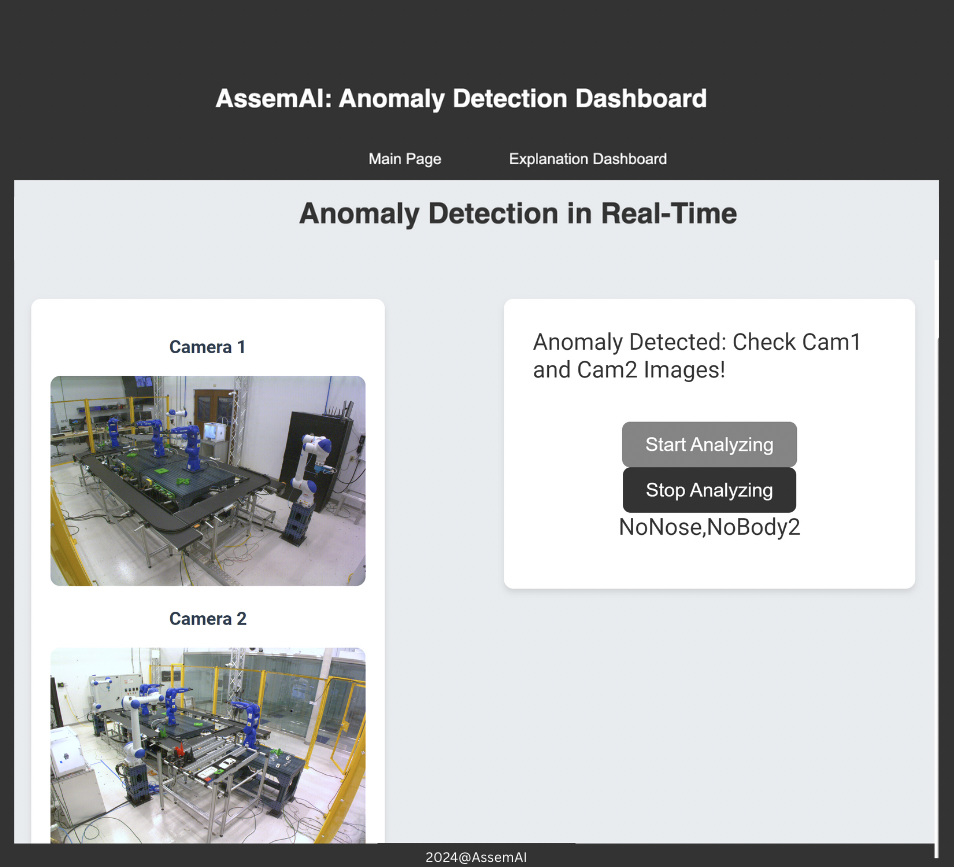}
  \caption{Web User interface of AssemAI}
  \label{fig:webui}
\end{figure}

\section{Deployment of AssemAI}
After training and testing the model, we deploy AssemAI on the Future Factories lab. This deployment setup can be seen in Figure \ref{fig:DepArch}. AssemAI requires two separate types of inputs to operate in real time. The script that utilizes the model should be able to obtain the current cycle state of the assembly process. This data tag allows the script to ensure that the images captured and input into the model are from cycle states four and nine. This tag is obtained by connecting to the OPC UA server running on the Programmable Logic Controller (PLC) and constantly reading the tag as it is updated. The cameras used in this deployment are separate from the PLC’s network. Images taken by the cameras can be acquired by connecting directly to them through a USB 3.0 wired connection. These cameras are industry-grade Basler cameras with a Python library, pyplon, which simplifies the image-capturing process. Since this is a wired connection, we can ensure that there is minimal communication lag time between the image request and the capturing process. As such, once the cycle state tag is read as either four or nine, the images are captured from the cameras and sent into the model for detection. Figure \ref{fig:webui} illustrates the web user interface of the deployed AssemAI.

\section{Conclusion and Further Work}
In this study, we derived an industry-standard dataset tailored for assembly processes using the novel YOLO-FF model and introduced AssemAI, a standardized image-based anomaly detection pipeline. We began with a simple CNN Model, which provided a baseline for anomaly detection through a straightforward architecture and standard training techniques. Building on this, we explored ViT and EfficientNet models to enhance classification accuracy and efficiency. A significant aspect of our approach is the focus on interpretability, where we aimed to understand both model and user lever explainability to high-level phenomena such as structural integrity. This not only improved model understanding but also provided actionable insights for anomaly detection. Our findings indicate that EfficientNet offers significant improvements over traditional methods. 
Future work should explore hybrid architectures, which take other modalities like time series and textual, further interpretability techniques and deployment on edge devices for real-time anomaly detection applications to extend these findings to other domains and production environments. Also, to improve how domain experts understand our approach, we suggest creating abstract representations of causal factors associated with anomalies. This approach focuses on linking sensor data to broader concepts like structural issues or gripper malfunctions, providing a more comprehensive view of anomalous events.

\section*{Acknowledgment}
This work is supported in part by NSF grants \#2119654, “RII Track 2 FEC: Enabling Factory to Factory (F2F) Networking for Future Manufacturing” and SCRA grant “Enabling Factory to Factory (F2F) Networking for Future Manufacturing across South Carolina”.




\bibliographystyle{IEEEtran}
\bibliography{IEEEabrv,IEEEexample}





\end{document}